# Emergent Behaviors from Folksonomy Driven Interactions

Massimiliano Dal Mas

me @ maxdalmas.com

## Abstract

To reflect the evolving knowledge on the Web this paper considers ontologies based on folksonomies according to a new concept structure called "Folksodriven" to represent folksonomies. This paper describes a research program for studying Folksodriven tags interactions leading to Folksodriven cluster behavior. The goal of the research is to understand the type of simple local interactions which produce complex and purposive group behaviors on Folksodriven tags.

We describe a synthetic, bottom-up approach to studying group behavior, consisting of designing and testing a variety of social interactions and cultural scenarios with Folksodriven tags. We propose a set of basic interactions which can be used to structure and simplify the process of both designing and analyzing emergent group behaviors. The presented behavior repertories was developed and tested on a folksonomy environment.

## Introduction

Intelligent behavior cannot be understood in isolation but in collaboration and competition of many individuals. The type of shared or group intelligence appears in consensus decision making in living systems. In the present work we consider the culture of a society defined by local rules of interaction. Concretely, to reflect the evolving knowledge this paper considers ontologies based on folksonomies according to a new concept structure called "Folksodriven" to represent folksonomies (*FD-tags*). Based on that we consider a knowledge sketched by folksonomy tags defined by the described *FD-tag*'s (Folksonomy Driven tag) behavior that is influenced by the *FD-tags* around it.

Imitation is a ubiquitous learning system in nature [1] being innate the propensity for imitation in nature. The ability to "imitate" can be used to explain the propagation on local *FD-tags* interactions both temporally and spatially as "social learning".

The background of related technologies are briefly described in the next section. Then, a system overview is presented, focusing on the "Folksodriven notation" and the "elasto-adaptative-dynamics" ontology matching to support the "Emergent Bahavior" for Folksodriven Clusters. Then, we validate our approach performing experiments on a sequence of tests. Lastly, some conclusions and future work are introduced.

---

## Related work

Artificial agents and their interactions have been studied extensively especially for robot tasks. A simple simulation for robots based on the principles of self organization is described in [2]. A report on asset of simulation with robots is described by [3]. While [4, 5] study the developing autonomous agents through learning.

A basic approach on simple navigation behaviors for multiple physical agents in described in [6].

Work on simulation ant colonies was studied for instance by [7, 8] to define the "Swarm Intelligence". While [9] describe a framework on incremental social learning on the collective problem-solving behavior of groups of animals and artificial agents. Artificial simulation of biologic organism has been studied extensively as in [10].

Respect the majority of Artificial Life works, this work considers the folksnomy tags in a driven environment(*FD-tags*) as agents modeling their "intelligence" on the elastic ontology matching described in [11-29] analyzing their behavior from the interaction among different *FD-tags*.

## Emergent Behavior

Emergent Behavior is a behavior of a system that is not explicitly described by the behavior of its parts but on their relationships to one another. It is a spontaneous creation of order, is presented all around us.

In Swarm Intelligence systems [7, 8] local interactions between agents leads to the emergence of global behavior: there is no centralized control structure. There is no centralized global structure everything is locally organized. Life systems are good examples because they manage complex tasks which we can appreciate – as the life in a colony of ants or bees, or birds flocking, animal herding, and fish schooling.

In this paper we will consider a particular type of collective behavior obtained empirically to be analyzed and used for the spontaneous arose on ontology matching among different folksonomy tags defined in a driven environment [11-29]. The next section introduces the framework used on folksonomy tags called "Folksodriven".

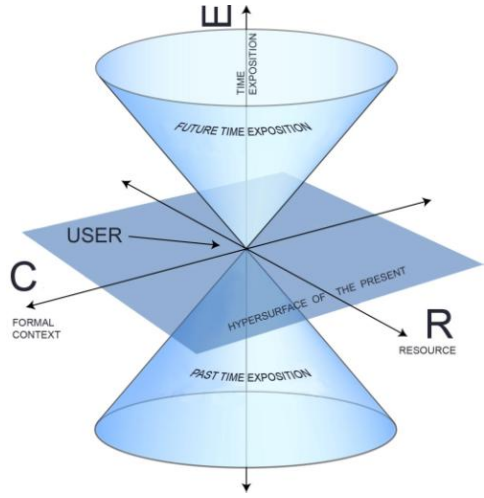

Fig. 1 **In a model of space-time, every point in space has four coordinates (*C*, *E*, *R*, *X*), representing a point in space and a precise moment in time *E***

## Folksodriven notation

In this section is introduced the formal notation used in the paper for the *Folksodriven tags* (*FD tags*) and the related *Time Exposition* (*E*).

- In a model of space-time (Fig. 1), every point in space has four coordinates ($x, y, z, t$), three of which represent a point in space, and the fourth a precise moment in time. Intuitively, each point represents an event that happened at a particular place at a precise moment.

The usage of the *four-vector* name assumes that its components refer to a "standard basis" on a Minkowski space [31]. Points in a Minkowski space are regarded as events in space-time. On a direction of time for the time vector we have:

- past directed time vector, whose first component is negative, to model the "history events" on the folksodriven notation
- future directed time vector, whose first component is positive, to model the "future events" on the folksodriven notation

A Folksodriven will be considered as a tuple (1) defined by finite sets composed by the *Formal Context* (*C*), the *Time Exposition (E)*, the *Resource (R)* and the ternary relation *X* – as in Fig. 1.

$$(1) \quad FD := (C, E, R, X)$$

As stated in [11-29] we consider a *Folksodriven tag* (*FD tag*) as a tuple (1) defined by finite sets composed by:

- *Formal Context (C)* is a triple $C:=(T, D, I)$ where the *Topic of Interest T* and the *Description D* are sets of data and *I* is a relation between *T* and *D*;
- *Time Exposition (E)* is the clickthrough rate (CTR) as the number of clicks on a *Resource (R)* divided by the number of times that the *Resource (R)* is displayed (impressions);
- *Resource (R)* is represented by the uri of the webpage that the user wants to correlate to a chosen tag;
- *X* is defined by the relation $X = C \times E \times R$ in a Minkowski vector space [31] delimited by the vectors *C*, *E* and *R*.

Following the above notation any text on a web page (*W*) can be depicted as a relation on the Folksodriven sets:

$$(2) \quad W(c,r) := \{(c,e,r) \in X \mid e \in E\}$$

## Folksodriven as a Network

On scientific literature folksonomies are modelled in order to obtain a number of model patterns and mathematical rules [32, 33]. The theoretical starting point considered is the emerging scientific network theory that aims to demonstrate mathematically that all complex networks follow a number of general regularities. The connection between the different vertices in the network forms a pattern. The pattern of all complex systems is the power law that arises because the network expands with the addition of new vertices which must be connected to other vertices in the network. A network with a degree distribution that follows a power law is a "scale-free network". An important characteristic of scale-free networks is the Clustering Coefficient distribution, which decreases as the node degree increases following a power law [32].

We consider a Folksodriven network in which nodes are Folksodriven tags (*FD tags*) and links are semantic acquaintance relationships between them according to the SUMO (http://www.ontologyportal.org) formal ontology that has been mapped to the WordNet lexicon (http://wordnet.princeton.edu). It is easy to see that *FD tags* tend to form groups, i.e. small groups in which tags are close related to each one, so we can think of such groups as a complete graph. In addition, the *FD tags* of a group also have a few acquaintance relationships to *FD tags* outside that group. Some *FD tags*, however, are so related to other tags (e.g.: workers, engineers) that are connected to a large number of groups. Those *FD tags* may be considered the hubs responsible for making such network a scale-free network [32]. While the disambiguation on *FD tags* can be done using the relations with other *FD tags*. In a scale-free network most nodes of a graph are not neighbors of one another but can be reached from every other by a small number of hops or steps, considering the mutual acquaintance of *FD tags*.

The network structure of "Folksodriven tags" (*FD tags*) – Folksodriven Structure Network (FSN) – was thought as a "Folsksonomy tags suggestions" for the user on a dataset built on chosen websites [11, 12].

The next section introduces the basic principles for the methodology used in this paper.

## Principles convention

The following principles are used as base for the methodology proposed in this work.

*FD-tag* is considered as an "agent" in Artificial Intelligence literature [34]. As tag it is considered homogeneous in its data dimensions.

*FD-tag* doesn't use any kind of one-to-one communication. All communication among FD-tags is based only on the Elastic Adaptive Ontology Matching described in [12]. That dynamical approach to the ontology matching can be considered as a stigmergic communication [1] for ubiquitous systems as observed in social behavior of insects. We consider here the analogy to cues bees use to indirectly exchange information.

*FD-tags* do not engage in any kind of explicit cooperation. While cooperation is implicit and is considered for the definition of merging ontologies. That cooperation occurs through the defined environment instead a direct communication. *FD-tags* affect each other by means of their external state considered as in [12].

*FD-tags* are dumb, they do not have hidden goals and they don't have a global vision of the environment. The common goal of all *FD-tags* is on define relations among other *FD-tags* to define a common ontology.

*FD-tags* can be linked each other according to [12]. A necessary condition for intelligent collective behavior is based on the ability to categorize the "object" – here considered as *FD-tags* – in an environment as similar or different respect them. Based on this ability it's possible to produce collective behavior as depicted in the following sections inspired by the biological life.

## Avoidance and Merging strategies

Finding a general schema on the merging/separation strategy for a *FD-tag* in a dynamic environment is not an easy task. Moreover the problem can become hard when considering an environment composed by a huge number of *FD-tags*.

To solve that problem knowledge on living systems can help us. From observation on insects and animals it was deducted how they do not have precise routines to avoid each other. Instead they use a simple behavior that is efficient most of the time. From their simple strategy and considering the ontology of reference we can depict the "Avoidance" behavior as:

```
Avoiding Other FD-tags:
If another FD-tag is not simile in Ontology(N)
  Avoid it
  otherwise Merge it
```

The Avoid/Merge *FD-tags* behavior works because all the *FD-tags* use as reference the same Ontology. In case a *FD-tag* fails to recognize another *FD-tag* respect the Ontology used as reference it will treat is a generic tag to avoid, using the following:

```
Avoiding Everything Else:
If another FD-tag is not in the Ontology(1)
  Avoid it, go.
If another FD-tag is not in the Ontology(2)
  Avoid it, go.
...
...
If another FD-tag is not in the Ontology(N)
  Avoid it, go.
```

This simple strategy has been effective in the majority of typical matching observed.

Merging can be implemented as inverse of *FD-tags* avoidance

```
Merging Other FD-tags:
If another FD-tag is simile in Ontology(N)
  Merge it
  otherwise Avoid it
Merging Everything Else:
If another FD-tag is in the Ontology(1)
  Merge it, go.
If another FD-tag is in the Ontology(2)
  Merge it, go.
...
...
If another FD-tag is in the Ontology(N)
  Merge it, go.
```

All conditions above area direct complement of those depicted in `Everything Else` behavior.

A corresponding biological system for merging can be considered on the osmotropotaxis of ants [1] that use the differential intensity of pheromone to decide the direction to take.

## FD tags Interaction

**Dispersion** can be considered as an ongoing process that maintains a desired distance between *FD tags* while they are related to other kind of *FD tags*.

The algorithm computes the local distribution (as the localization of other *FD tags*) in order to decide in which direction to move. The algorithm computes the local centroid to determine where most of the nearby *FD tags* are, and then moves away from that direction. In situation of high density of *FD tags*, the system can take a long time to achieve a dispersed state since local interactions

---

[1] Stigmergic communication is based on modifications to the environments determined by biological life instead of the direct message passing

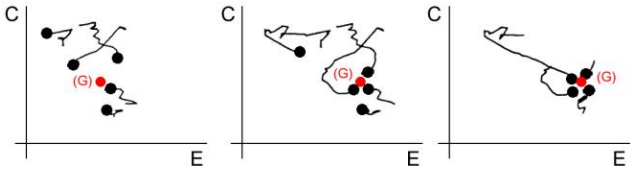

Fig. 2 **Goal-ing behavior of 5 *FD-tags* using *Time Exposition* (*E*) and *Formal Context* (*C*) as coordinate system. Started in an arbitrary initial configuration, four of the *FD-tags* reached the *goal FD-tags* (*G*) within 200 seconds, and the fifth joined them 50 seconds later**

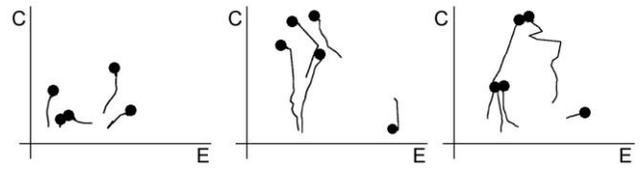

Fig. 3 **Flocking behavior of 5 *FD-tag*s using *Time Exposition* (*E*) and *Formal Context* (*C*) as coordinate system: one of the *FD-tag* diverges without affecting the behavior of the others and it is unable to rejoin the group of the FD-tags**

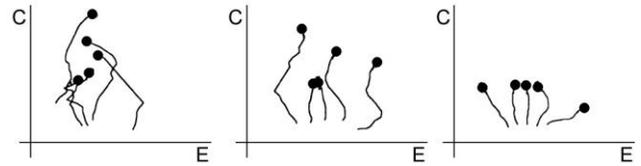

Fig. 4 **Flocking behavior of 5 *FD-tags* using *Time Exposition* (*E*) and *Formal Context* (*C*) as coordinate system: the FD-tag is coherent in the group of the FD-tags. The group moves and stay together**

propagate far, and the motion of individual *FD tags* can disturb the state of many others.

```
If another FD-tag is sensed
Within the "personal space"
  do no link to it
  then go forward for a fixed time.
If multiple FD-tags are sensed
  move away from the local centroid,
  then go forward for a fixed time.
Otherwise, stop.
```

**Aggregation** is the inverse of dispersion. Its goal is to bring all of the *FD-tags* within a determined distance from each other.

**Dispersing** and **aggregation** are similar to behaviors exhibited by army ants in the process of maintain stable the temperature of an anthill. Individual ants aggregate and disperse by following the local temperature gradient [35]. The temperature control is used by ants for their interactions. Similarly, elasticity could be used as a means of collecting and distributing *FD-tags* in the environment [12], as in the avoidance and merging behaviors described before.

The simplest **goal-ing** strategy can be a done in a greedy local way to link the *FD-tags* to the *FD-tag* considered as goal.

```
Link the FD-tag toward the goal FD-tag,
  go forward.
If linked at goal, stop.
```

In Fig. 2 the data illustrated that the actual trajectories are far from the optimal due to sensitivity in the elastic measuring.

As long as the density of *FD-tags* is sufficiently low it is performed an efficient individual goal-ing. If enough *FD-tags* are trying to goal-ing, they begin to link with each other. A worse situation happens if the *FD-tags* are subject to elastic sensitivity [12] and errors.

All of the above conditions are common in *FD-Tags*, suggesting the need for some form of group checking for the common value of elasticity [12].

**Flocking** is an essential means for a group of *FD-tags* to move together. It can be viewed as a combination of avoidance, following, aggregation, and dispersion. Each of these constituent behaviors produces an effect or command either telling the *FD-tag* to stop, go, or link. Flocking weighs and combines these outputs.

The choice of weights on the behavior outputs depends on the dynamics and mechanics of the *FD-tags*, and the range of the links. Due to the number of tunable parameters involved, flocking is the most complex basic interaction implemented in this work so far. It consists of a combination of *avoidance* and *aggregation* only. *Following* and *dispersion* resulted from the *FD-tags* combination. Avoidance has the highest priority; the strength of aggregation is proportionate to the *FD-tags* distance from the computed centroid.

The flocking algorithm was inspired the simulation of the bird flocking by [36]. However, the *FD-tag* implementation requires many more details that the simulation, due to the more complex dynamics. Fig. 3 and Fig. 4 demonstrate two runs of the flocking behavior with five *FD-tags*.

## Testing

In order to test the sufficiency and the additive properties of the basic interactions, we are currently implementing a number of more complex ontology matching among different kind of not-structured ontology defined by folksonomies. These compound behaviors emerge as by temporal sequences of basic interactions, each triggered by the appropriate conditions in the environment defined by different kind of folksonomy tags. For instance, a matching task is initiated by a dispersing behavior. The *FD-tags* disperse until they find link connections. Once a *FD-tag* is holding a link connection, it starts to link to the goal *FD-tag*. If it encounters a *FD-tag* with a different elastic ontology it avoids it. Instead, if a *FD-tag* is near a *FD-tag* with the same elastic it links to that. More than two following same elasticity *FD-tags* forma flock. Depending on the *FD-tag* considered as goal and the built-in linking to other *FD-tags*, *FD-tag* continue to indefinitely search other *FD-tags* to be linked, or if not finding any it stops to search other *FD-tags*. Combining linking temporally relies on the *FD-tags* to perceive the elasticity of *FD-tags* linking. Combining linking interaction is more challenging. If two groups of linked *FD-tags* come into contact, they must not create an unexpected higher order behavior.

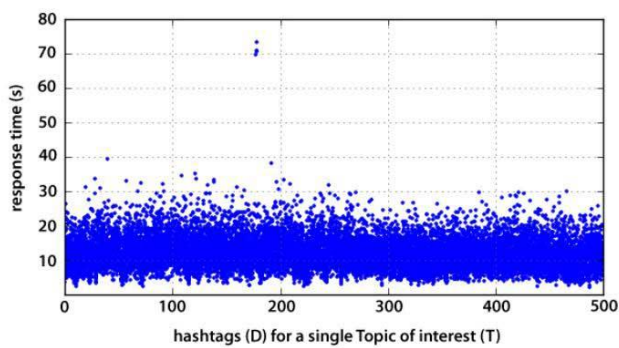

Fig. 5 **Distribution of response time obtained by issuing 500 *#hashtags* (*D*) for a single *Topic of interest* (*T*)**

## Experimental observation

For the experimental observations it was considered the data collection acquired analyzing the hashtags of Twitter [37].
Hashtags are an example of folksonomies for social networks; they are used to identify groups and topics (the short messages called Tweet in Twitter). Hashtags are neither registered nor controlled by any one user or group of users; they are used to identify groups and topics to identify short messages on microblogging social networking services such as: Twitter, Facebbok, Linkedin, YouTube, Instagram, Pintrest…
For the Experimental Observation it was considered only 100K tweets connected to "world news" topics.
The Folksodriven notation was used for the content analysis on dynamic ontology matching for the Twitter #hashtags correlated to a topic T (for the evaluation on five different topics chosen).
The system performance was measured in terms of efficiency of the analysis and matching process.
To measure the efficiency of the Elastic Adaptive Ontology Matching a stress test was done [13, 14, 19] performing the two most expensive tasks occurring at run-time:
- the time needed by the algorithm to analyze a new set of 500K #hashtags from Twitter to be deployed and to generate the *FD-tags*,
- the time needed to automatically generate matching for the *Folksodriven Structure Network* (*FSN*)
Fig. 5 shows the distribution of response time obtained by issuing 500 Twitter *#hashtags* (*D*) for a single *Topic of interest* (*T*).

## Conclusion

This paper has described a study on social interaction towards the folksonomy behavior. The goal of this work is to understand the simple local interactions among folksonomy tags that lead to ontology matching behavior.
The behavior in the Collective Intelligence is described with a button-up approach, consisting of designing and testing different kind of tag interactions on different ontology scenarios. A set of basic *FD-tags* interactions can be used to structure the process to define emergent group behaviors in a simply way.
To synthesizing the group behavior on *FD-tags* we look to the interaction on biological systems.

**Massimiliano Dal Mas** is an engineer working on webservices and is interested in knowledge engineering. His interests include: user interfaces and visualization for information retrieval, automated Web interface evaluation and text analysis, empirical computational linguistics, text data mining, knowledge engineering and artificial intelligence. He received BA, MS degrees in Computer Science Engineering from the Politecnico di Milano, Italy. He won the thirteenth edition 2008 of the CEI Award for the "best degree thesis" with a dissertation on "Semantic technologies for industrial purposes" (Supervisor Prof. M. Colombetti). In 2012, he received the "best paper award" at the IEEE Computer Society Conference on Evolving and Adaptive Intelligent System (EAIS 2012) at Carlos III University of Madrid, Madrid, Spain. In 2013, he received the "best paper award" at the ACM Conference on Web Intelligence, Mining and Semantics (WIMS 2013) at Universidad Autónoma de Madrid, Madrid, Spain. His paper at 2013 W3C Workshop on Publishing using CSS3 & HTML5 has been appointed as "position paper", Paris, France.